\documentclass{article} 
\usepackage{nips14submit_e,times,graphicx}
\usepackage{url}
\usepackage{amsmath}
\usepackage{amssymb}

\title{A Simple Way to Initialize Recurrent  Networks of Rectified Linear Units}

\author{ Quoc V. Le, Navdeep Jaitly, Geoffrey E.
  Hinton\\ Google\\ 
  \\ }

\nipsfinalcopy 

\begin{document}

\maketitle

\begin{abstract}
  Learning long term dependencies in recurrent networks is difficult
  due to vanishing and exploding gradients.  To overcome this
  difficulty, researchers have developed sophisticated optimization
  techniques and network architectures. In this paper, we propose a
  simpler solution that use recurrent neural networks composed of
  rectified linear units.  Key to our solution is the use of the
  identity matrix or its scaled version to initialize the recurrent
  weight matrix. We find that our solution is comparable to a standard
  implementation of LSTMs on our four benchmarks: two toy problems
  involving long-range temporal structures, a large language modeling
  problem and a benchmark speech recognition problem.
\end{abstract}

\section{Introduction}

Recurrent neural networks (RNNs) are very powerful dynamical systems
and they are the natural way of using neural networks to map an input
sequence to an output sequence, as in speech recognition and machine
translation, or to predict the next term in a sequence, as in language
modeling.  However, training RNNs by using back-propagation through
time~\cite{rumelhart1986learning} to compute error-derivatives can be
difficult. Early attempts suffered from vanishing and exploding
gradients~\cite{hoch01} and this meant that they had great difficulty
learning long-term dependencies.  Many different methods have been
proposed for overcoming this difficulty.


A method that has produced some impressive
results~\cite{martens2010deep,martens11} is to abandon stochastic
gradient descent in favor of a much more sophisticated Hessian-Free
(HF) optimization method.  HF operates on large mini-batches and is
able to detect promising directions in the weight-space that have very
small gradients but even smaller curvature. Subsequent work, however,
suggested that similar results could be achieved by using stochastic
gradient descent with momentum provided the weights were initialized
carefully~\cite{sutskever2013importance} and large gradients were
clipped~\cite{pascanu2012difficulty}.  Further developments of the HF
approach look promising~\cite{sutskever2011generating,martens12} but
are much harder to implement than popular simple methods such as
stochastic gradient descent with
momentum~\cite{sutskever2013importance} or adaptive learning rates for
each weight that depend on the history of its
gradients~\cite{duchi2011adaptive,hinton12b}.

The most successful technique to date is the Long Short Term Memory
(LSTM) Recurrent Neural Network which uses stochastic gradient
descent, but changes the hidden units in such a way that the
backpropagated gradients are much better
behaved~\cite{hochreiter97}. LSTM replaces logistic or tanh hidden
units with ``memory cells'' that can store an analog value.  Each
memory cell has its own input and output gates that control when
inputs are allowed to add to the stored analog value and when this
value is allowed to influence the output. These gates are logistic
units with their own learned weights on connections coming from the
input and also the memory cells at the previous time-step. There is
also a forget gate with learned weights that controls the rate at
which the analog value stored in the memory cell decays.  For periods
when the input and output gates are off and the forget gate is not
causing decay, a memory cell simply holds its value over time so the
gradient of the error {\it w.r.t.} its stored value stays constant
when backpropagated over those periods.

The first major success of LSTMs was for the task of unconstrained
handwriting recognition~\cite{graves2009novel}. Since then, they have
achieved impressive results on many other tasks including speech
recognition~\cite{graves2013speech,graves2014towards}, handwriting
generation~\cite{graves2013generating}, sequence to sequence
mapping~\cite{sutskever14}, machine
translation~\cite{luong2014addressing,bahdanau2014neural}, image
captioning~\cite{vinyals2014show,kiros2014unifying},
parsing~\cite{vinyals2014grammar} and predicting the outputs of simple
computer programs~\cite{zaremba2014learning}.

The impressive results achieved using LSTMs make it important to
discover which aspects of the rather complicated architecture are
crucial for its success and which are mere passengers. It seems
unlikely that Hochreiter and Schmidhuber's~\cite{hochreiter97} initial
design combined with the subsequent introduction of forget
gates~\cite{gers2000learning,gers2003learning} is the optimal design:
at the time, the important issue was to find any scheme that could
learn long-range dependencies rather than to find the minimal or
optimal scheme.  One aim of this paper is to cast light on what
aspects of the design are responsible for the success of LSTMs.

Recent research on deep feedforward networks has also produced some
impressive results~\cite{kriz12, dahl12b} and there is now a consensus
that for deep networks, rectified linear units (ReLUs) are easier to
train than the logistic or tanh units that were used for many
years~\cite{nair10,zeiler2013rectified}.  At first sight, ReLUs seem
inappropriate for RNNs because they can have very large outputs so
they might be expected to be far more likely to explode than units
that have bounded values. A second aim of this paper is to explore
whether ReLUs can be made to work well in RNNs and whether the ease of
optimizing them in feedforward nets transfers to RNNs.


\section{The initialization trick}

In this paper, we demonstrate that, with the right initialization of
the weights, RNNs composed of rectified linear units are relatively
easy to train and are good at modeling long-range dependencies. The
RNNs are trained by using backpropagation through time to get
error-derivatives for the weights and by updating the weights after
each small mini-batch of sequences. Their performance on test data is
comparable with LSTMs, both for toy problems involving very long-range
temporal structures and for real tasks like predicting the next word in
a very large corpus of text.

We initialize the recurrent weight matrix to be the identity matrix
and biases to be zero. This means that each new hidden state vector is
obtained by simply copying the previous hidden vector then adding on
the effect of the current inputs and replacing all negative states by
zero. In the absence of input, an RNN that is composed of ReLUs and
initialized with the identity matrix (which we call an IRNN) just
stays in the same state indefinitely. The identity initialization has
the very desirable property that when the error derivatives for the
hidden units are backpropagated through time they remain constant
provided no extra error-derivatives are added.  This is the same
behavior as LSTMs when their forget gates are set  so
that there is no decay and it makes it easy to learn very long-range
temporal dependencies.

We also find that for tasks that exhibit less long range dependencies,
scaling the identity matrix by a small scalar is an effective
mechanism to forget long range effects. This is the same behavior as
LTSMs when their forget gates are set so that the memory decays fast.

Our initialization scheme bears some resemblance to the idea of
Mikolov et al.~\cite{mikolov2014learning}, where a part of the weight
matrix is fixed to identity or approximate identity. The main
difference of their work to ours is the fact that our network uses the
rectified linear units and the identity matrix is only used for
initialization. The scaled identity initialization was also proposed
in Socher et al.~\cite{socher13parsing} in the context of
tree-structured networks but without the use of ReLUs. Our work is
also related to the work of Saxe et al.~\cite{saxe2013exact}, who
study the use of orthogonal matrices as initialization in deep
networks.

\section{Overview of the experiments}

Consider a recurrent net with two input units. At each time step, the
first input unit has a real value and the second input unit has a
value of 0 or 1 as shown in figure \ref{fig:add}.  The task is to
report the sum of the two real values that are marked by having a 1 as
the second input~\cite{hochreiter97,hoch01,martens11}. IRNNs can learn
to handle sequences with a length of 300, which is a challenging
regime for other algorithms.

Another challenging toy problem is to learn to classify the MNIST
digits when the 784 pixels are presented sequentially to the recurrent
net. Again, the IRNN was better than the LSTM, having been able to
achieve 3\% test set error compared to 34\% for LSTM.

While it is possible that a better tuned LSTM (with a different
architecture or the size of the hidden state) would outperform the
IRNN for the above two tasks, the fact that the IRNN performs as well
as it does, with so little tuning, is very encouraging, especially
given how much simpler the model is, compared to the LSTM.


We also compared IRNNs with LSTMs on a large language modeling
task. Each memory cell of an LSTM is considerably more complicated
than a rectified linear unit and has many more parameters, so it is
not entirely obvious what to compare.  We tried to balance for both
the number of parameters and the complexity of the architecture by
comparing an LSTM with N memory cells with an IRNN with four layers of
N hidden units, and an IRNN with one layer and 2N hidden units. Here we find
that the IRNN gives results comparable to the equivalent LSTM.

Finally, we benchmarked IRNNs and LSTMs on a acoustic modeling task on
TIMIT. As the tasks only require a short term memory of the inputs, we
used a the identity matrix scaled by 0.01 as initialization for the
recurrent matrix. Results show that our method is also comparable to
LSTMs, despite being a lot simpler to implement.

\section{Experiments}
In the following experiments, we compare IRNNs against LSTMs, RNNs
that use tanh units and RNNs that use ReLUs with random Gaussian
initialization.


For IRNNs, in addition to the recurrent weights being initialized at
identity, the non-recurrent weights are initialized with a random
matrix, whose entries are sampled from a Gaussian distribution with
mean of zero and standard deviation of 0.001.

Our implementation of the LSTMs is rather standard and includes the
forget gate. It is observed that setting a higher initial forget gate
bias for LSTMs can give better results for long term dependency
problems. We therefore also performed a grid search for the initial
forget gate bias in LSTMs from the set $\{1.0, 4.0, 10.0,
20.0\}$. Other than that we did not tune the LTSMs much and it is
possible that the results of LSTMs in the experiments can be improved.

In addition to LSTMs, two other candidates for comparison are RNNs
that use the tanh activation function and RNNs that use ReLUs with
standard random Gaussian initialization. We experimented with several
values of standard deviation for the random initialization Gaussian
matrix and found that values suggested in~\cite{sussillo2015} work
well.

To train these models, we use stochastic gradient descent with a fixed
learning rate and gradient clipping. To ensure that good
hyperparameters are used, we performed a grid search over several
learning rates $\alpha = \{10^{-9},10^{-8},...,10^{-1}\}$ and gradient
clipping values $gc = \{1, 10, 100,
1000\}$~\cite{graves13c,sutskever14}. The reported result is the best
result over the grid search. We also use the same batch size of 16
examples for all methods. The experiments are carried out using the
DistBelief infrastructure, where each experiment only uses one
replica~\cite{le12,dean12}.


\subsection{The Adding Problem}
The adding problem is a toy task, designed to examine the power of
recurrent models in learning long-term
dependencies~\cite{hochreiter97,hoch01}. This is a sequence regression
problem where the target is a sum of two numbers selected in a
sequence of random signals, which are sampled from a uniform
distribution in [0,1]. At every time step, the input consists of a
random signal and a mask signal. The mask signal has a value of zero
at all time steps except for two steps when it has values of 1 to
indicate which two numbers should be added. An example of the adding
problem is shown in figure~\ref{fig:add} below.

\begin{figure}[h!]
\centering
\includegraphics[width=0.6\textwidth]{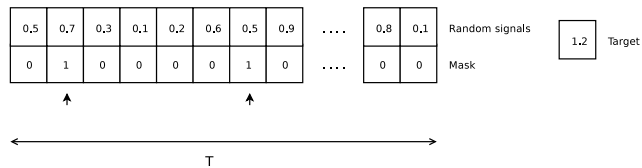}
\caption{An example of the ``adding'' problem, where the target is 1.2
  which is the sum of 2nd and the 7th numbers in the first
  sequence~\cite{martens11}.}
\label{fig:add}
\end{figure}

A basic baseline is to always predict the sum to have a value of 1
regardless of the inputs. This will give the Mean Squared Error (MSE)
around 0.1767. The goal is to train a model that achieves MSE well
below 0.1767.

The problem gets harder as the length of the sequence $T$ increases
because the dependency between the output and the relevant inputs
becomes more remote. To solve this problem, the recurrent net must
remember the first number or the sum of the two numbers accurately
whilst ignoring all of the irrelevant numbers.

We generated a training set of 100,000 examples and a test set of
10,000 examples as we varied $T$.  We fixed the hidden states to have
$100$ units for all of our networks (LSTMs, RNNs and IRNNs). This
means the LSTMs had more parameters by a factor of about 4 and also
took about 4 times as much computation per timestep.

\begin{figure}[h!]
\centering
\includegraphics[width=0.45\textwidth]{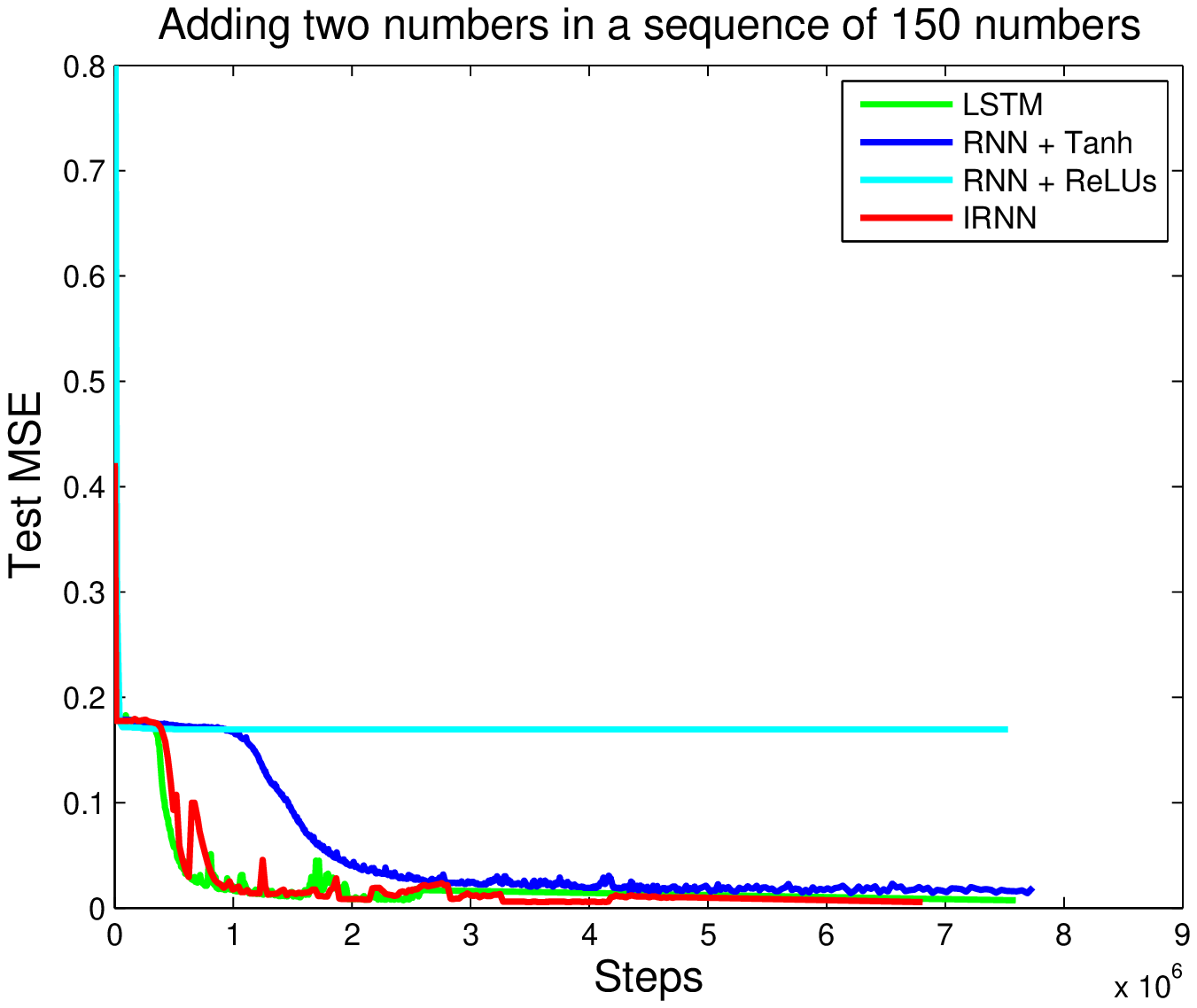}
\includegraphics[width=0.45\textwidth]{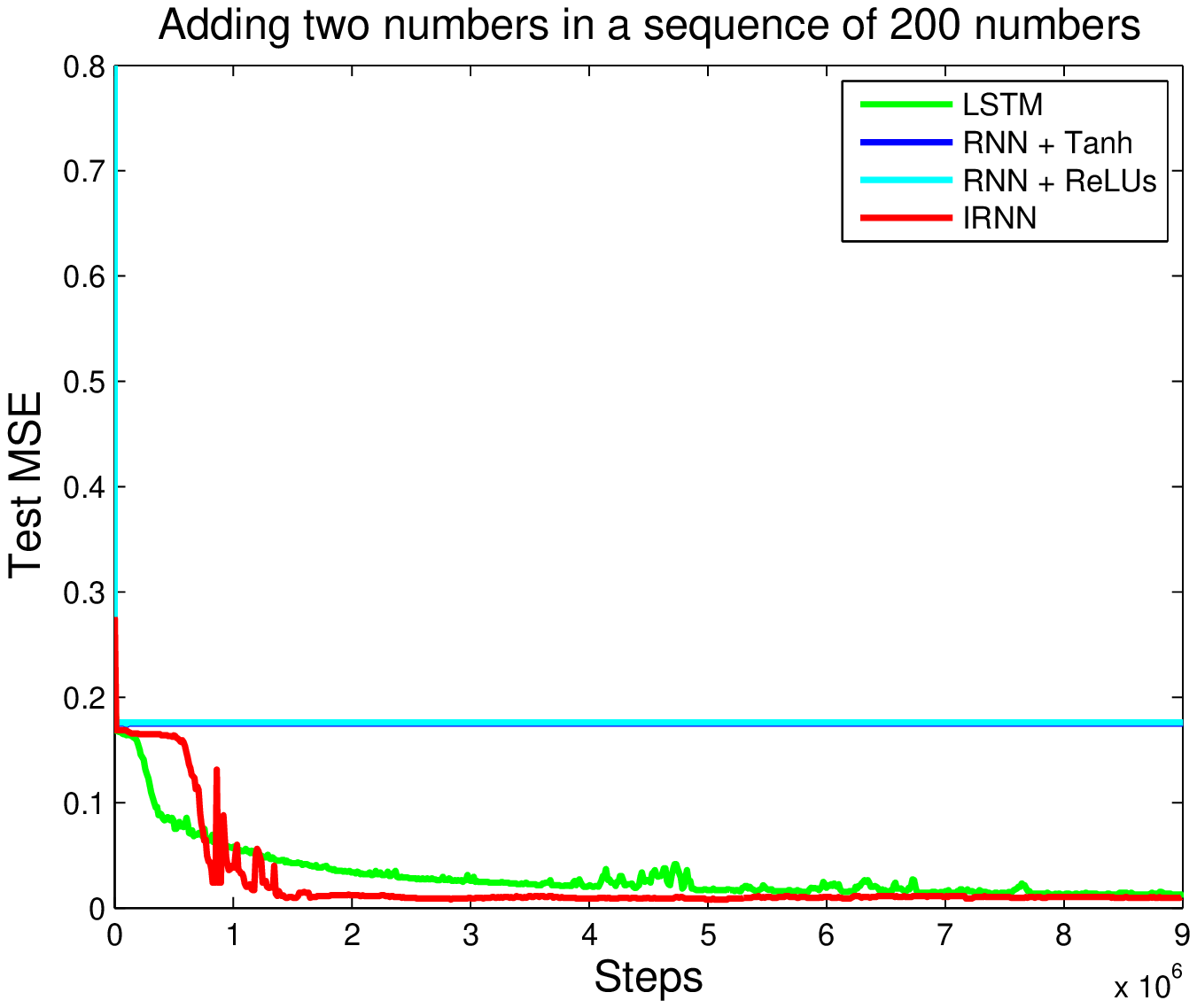}
\includegraphics[width=0.45\textwidth]{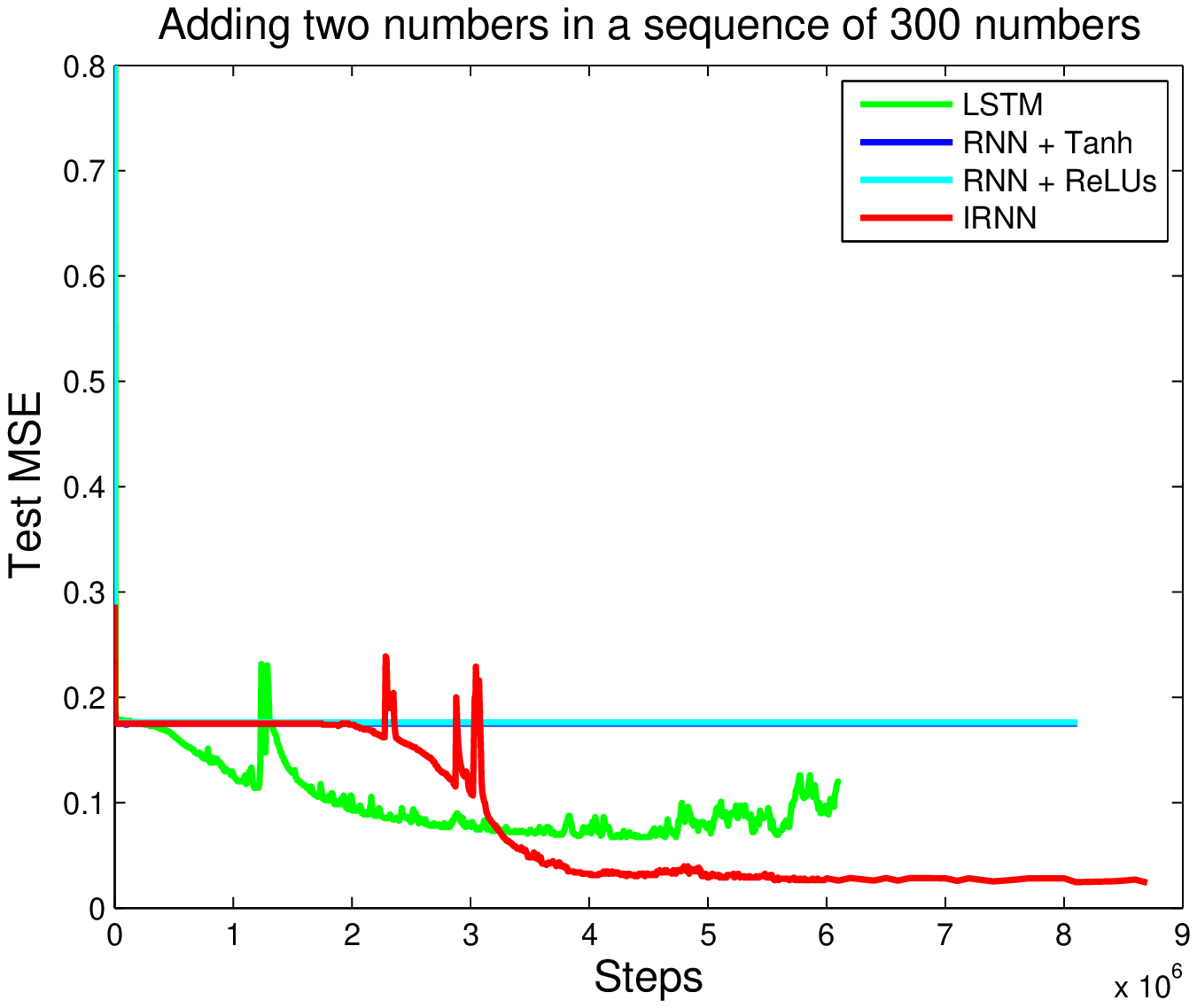}
\includegraphics[width=0.45\textwidth]{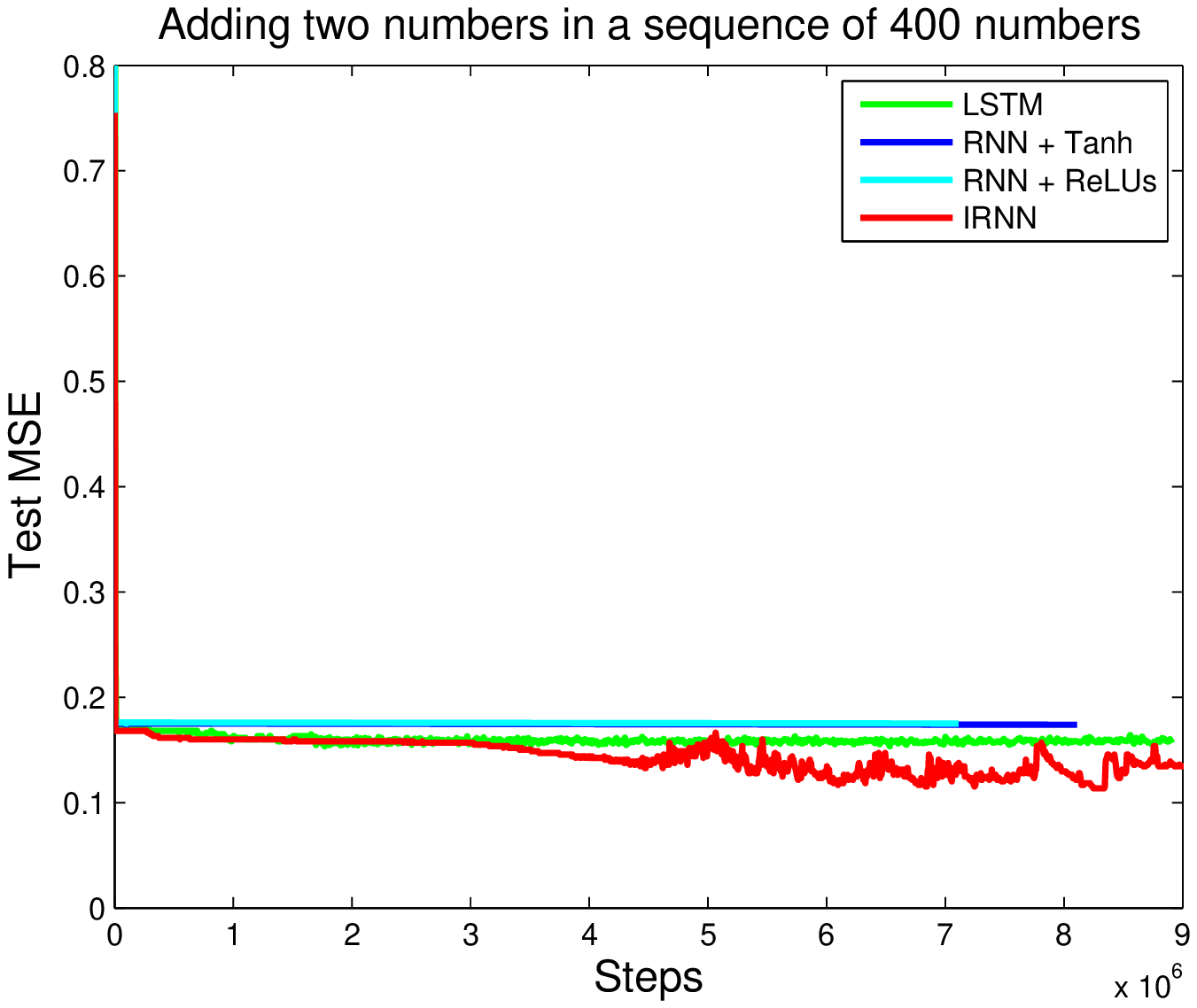}
\caption{The results of recurrent methods on the ``adding'' problem
  for the case of $T=150$ (top left), $T = 200$ (top right), $T = 300$
  (bottom left) and $T = 400$ (bottom right). The objective function
  is the Root Mean Squared Error, reported on the test set of 10,000
  examples. Note that always predicting the sum to be 1 should give
  MSE of 0.1767.}
\label{fig:adding}
\end{figure}

As we varied $T$, we noticed that both LSTMs and RNNs started to
struggle when $T$ is around 150. We therefore focused on investigating
the behaviors of all models from this point onwards. The results of
the experiments with $T = 150, T = 200, T = 300, T = 400$ are reported
in figure~\ref{fig:adding} below (best hyperparameters found during
grid search are listed in table~\ref{tab:adding-problem-hyp}).

\begin{table}[h!]
\begin{small}
\centering
\begin{tabular}{|l|l|l|l|}
\hline
{\bf T}           &  {\bf LSTM}                 &  {\bf RNN + Tanh} & {\bf IRNN} \\\hline \hline
150               &  $lr=0.01, gc=10, fb=1.0$   &   $lr=0.01, gc=100$ & $lr=0.01, gc=100$       \\\hline 
200               &  $lr=0.001, gc=100, fb=4.0$ &   N/A                  & $lr=0.01, gc=1$       \\\hline 
300               &  $lr=0.01, gc=1, fb=4.0$    &   N/A                  & $lr=0.01, gc=10$       \\\hline 
400               &  $lr=0.01, gc=100, fb=10.0$ &   N/A                  & $lr=0.01, gc=1$       \\\hline 
\end{tabular}
\caption{Best hyperparameters found for adding problems after grid
  search. $lr$ is the learning rate, $gc$ is gradient clipping, and
  $fb$ is forget gate bias. N/A is when there is no hyperparameter
  combination that gives good result.}
\label{tab:adding-problem-hyp}
\end{small}
\end{table}

The results show that the convergence of IRNNs is as good as
LSTMs. This is given that each LSTM step is more expensive than an
IRNN step (at least 4x more expensive). Adding two numbers in a
sequence of 400 numbers is somewhat challenging for both algorithms.

\subsection{MNIST Classification from a Sequence of Pixels}

Another challenging toy problem is to learn to classify the MNIST
digits~\cite{lecun98} when the 784 pixels are presented sequentially
to the recurrent net. In our experiments, the networks read one pixel
at a time in scanline order ({\it i.e.}  starting at the top left
corner of the image, and ending at the bottom right corner). The
networks are asked to predict the category of the MNIST image only
after seeing all 784 pixels. This is therefore a huge long range
dependency problem because each recurrent network has 784 time steps.

To make the task even harder, we also used a fixed random permutation
of the pixels of the MNIST digits and repeated the experiments.

\begin{figure}[htb]
\centering
\includegraphics[width=0.49\textwidth]{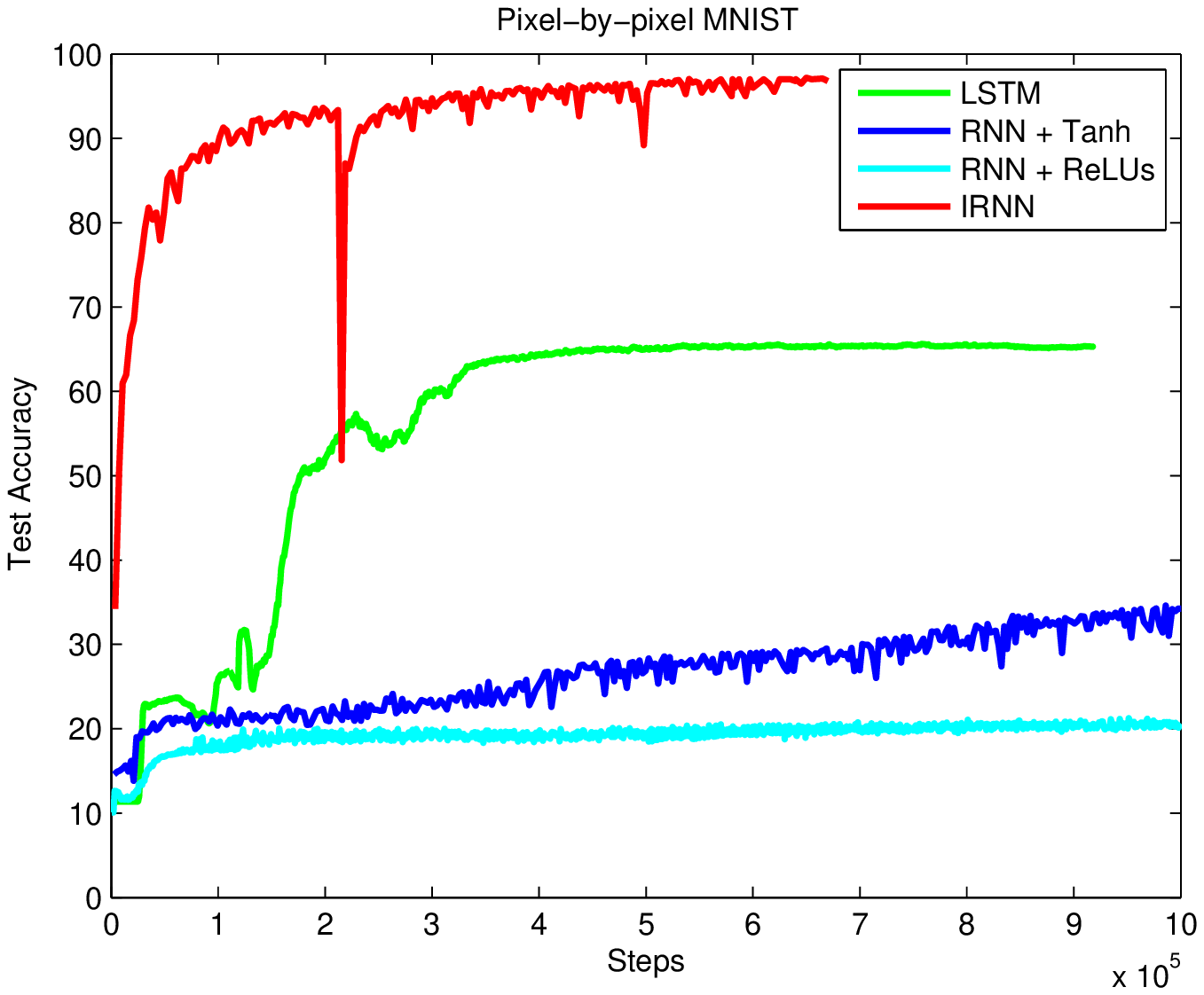}
\includegraphics[width=0.49\textwidth]{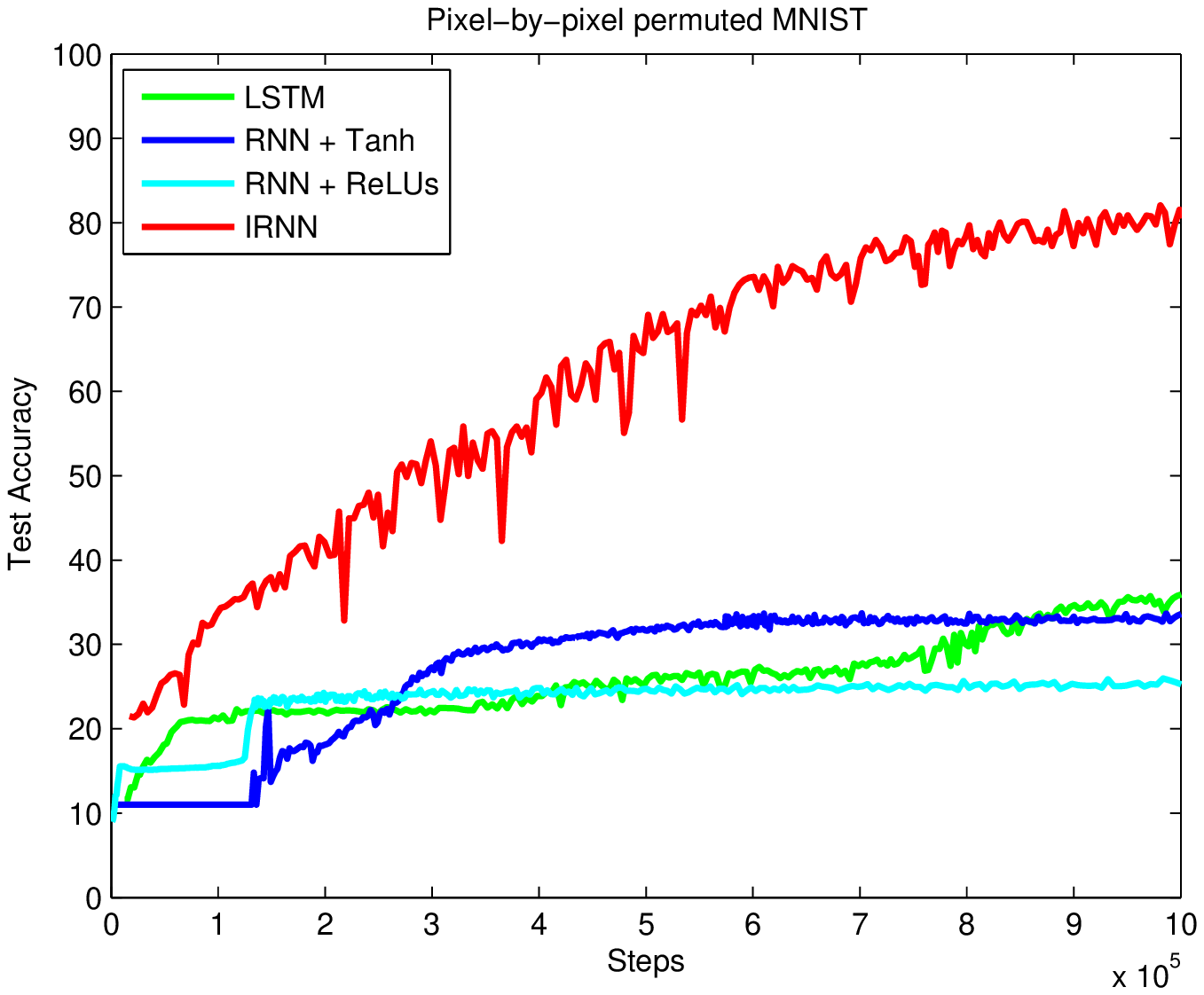}
\caption{The results of recurrent methods on the ``pixel-by-pixel
  MNIST'' problem. We report the test set accuracy for all
  methods. Left: normal MNIST. Right: permuted MNIST.}
\label{fig:mnist}
\end{figure}

All networks have 100 recurrent hidden units. We stop the optimization
after it converges or when it reaches 1,000,000 iterations and report
the results in figure~\ref{fig:mnist} (best hyperparameters are listed
in table~\ref{tab:mnist-problem-hyp}).

\begin{table}[h!]
\begin{small}
\centering
\begin{tabular}{|l|l|l|l|l|}
\hline
{\bf Problem}       &  {\bf LSTM} & {\bf RNN + Tanh} & {\bf RNN + ReLUs} & {\bf IRNN} \\\hline \hline
MNIST               &  $lr=0.01, gc=1$ & $lr=10^{-8}, gc=10$ & $lr=10^{-8}, gc=10$ & $lr=10^{-8}, gc=1$       \\
                    &  $fb=1.0$        &                    &                    &                         \\\hline                    
permuted       &  $lr=0.01, gc=1$ & $lr=10^{-8}, gc=1$ & $lr=10^{-6}, gc=10$  & $lr=10^{-9}, gc=1$       \\
MNIST                    &  $fb=1.0$        &                    &                    &                         \\\hline
\end{tabular}
\caption{Best hyperparameters found for pixel-by-pixel MNIST problems after grid
  search. $lr$ is the learning rate, $gc$ is gradient clipping, and
  $fb$ is the forget gate bias.}
\label{tab:mnist-problem-hyp}
\end{small}
\end{table}

The results using the standard scanline ordering of the pixels show
that this problem is so difficult that standard RNNs fail to work,
even with ReLUs, whereas the IRNN achieves 3\% test error rate which
is better than most off-the-shelf linear
classifiers~\cite{lecun98}. We were surprised that the LSTM did not
work as well as IRNN given the various initialization schemes that we
tried. While it still possible that a better tuned LSTM would do
better, the fact that the IRNN perform well is encouraging.

Applying a fixed random permutation to the pixels makes the problem
even harder but IRNNs on the permuted pixels are still better
than LSTMs on the non-permuted pixels.

The low error rates of the IRNN suggest that the model can discover
long range correlations in the data while making weak assumptions
about the inputs. This could be important to have for problems when
input data are in the form of variable-sized vectors ({\it e.g.}  the
repeated field of a
protobuffer~\footnote{\url{https://code.google.com/p/protobuf/}}).

\subsection{Language Modeling}
We benchmarked RNNs, IRNNs and LSTMs on the one billion word language
modelling dataset~\cite{DBLP:journals/corr/ChelbaMSGBK13}, perhaps the
largest public benchmark in language modeling. We chose an output
vocabulary of 1,000,000 words.

As the dataset is large, we observed that the performance of recurrent
methods depends on the size of the hidden states: they perform better
as the size of the hidden states gets larger ({\it
  cf.}~\cite{DBLP:journals/corr/ChelbaMSGBK13}). We however focused on
a set of simple controlled experiments to understand how different
recurrent methods behave when they have a similar number of
parameters. We first ran an experiment where the number of hidden
units (or memory cells) in LSTM are chosen to be 512. The LSTM is
trained for 60 hours using 32 replicas. Our goal is then to check how
well IRNNs perform given the same experimental environment and
settings. As LSTM have more parameters per time step, we compared them
with an IRNN that had 4 layers and same number of hidden units per
layer (which gives approximately the same numbers of parameters).

We also experimented shallow RNNs and IRNNs with 1024 units. Since the
output vocabulary is large, we projected the 1024 hidden units to a
linear layer with 512 units before the softmax.  This avoids greatly
increasing the number of parameters.

The results are reported in table~\ref{tab:1b}, which show that the
performance of IRNNs is closer to the performance of LSTMs for this
large-scale task than it is to the performance of RNNs.

\begin{table}[h!]
\centering
\begin{tabular}{|l|c|c|}
\hline
{\bf Methods}         &  {\bf Test perplexity} \\\hline \hline
LSTM (512 units)   &  68.8            \\
IRNN (4 layers, 512 units)   &  69.4            \\
IRNN (1 layer, 1024 units + linear projection with 512 units before softmax)  &  70.2            \\
RNN (4 layer, 512 tanh units)  &  71.8            \\
RNN (1 layer, 1024 tanh units + linear projection with 512 units before softmax)  &  72.5            \\\hline 
\end{tabular}
\caption{Performances of recurrent methods on the 1 billion word
  benchmark.}
\label{tab:1b}
\end{table}

\subsection{Speech Recognition}
We performed Phoneme recognition experiments on TIMIT with IRNNs and
Bidirectional IRNNs and compared them to RNNs, LSTMs and Bidirectional
LSTMs and RNNs. Bidirectional LSTMs have been applied previously to TIMIT
in \cite{graves2013hybrid}. In these experiments we generated phoneme
alignments from Kaldi \cite{Kaldi} using the recipe reported in
\cite{jaitly2014exploring} and trained all RNNs with two and five
hidden layers. Each model was given log Mel filter bank spectra with
their delta and accelerations, where each frame was 120 (=40*3)
dimensional and trained to predict the phone state (1 of 180). Frame
error rates (FER) from this task are reported in
table~\ref{tab:speech}.

\begin{table}[h!]
\centering
\begin{tabular}{|l|c|c|}
\hline {\bf Methods} & {\bf Frame error rates (dev / test) } \\\hline
\hline RNN (500 neurons, 2 layers) & 35.0 / 36.2 \\ LSTM (250 cells, 2
layers) & 34.5 / 35.4 \\ iRNN (500 neurons, 2 layers) & 34.3 / 35.5
\\\hline \hline

RNN (500 neurons, 5 layers)                 &  35.6 / 37.0   \\
LSTM (250 cells, 5 layers)                  &  35.0 / 36.2   \\
iRNN (500 neurons, 5 layers)                &  33.0 / 33.8   \\\hline \hline 

Bidirectional RNN (500 neurons, 2 layers)   &  31.5 / 32.4   \\
Bidirectional LSTM (250 cells, 2 layers)    &  29.6 / 30.6   \\
Bidirectional iRNN (500 neurons, 2 layers)  &  31.9 / 33.2        \\\hline \hline 

Bidirectional RNN (500 neurons, 5 layers)   &  33.9 / 34.8   \\
Bidirectional LSTM (250 cells, 5 layers)    &   28.5 / 29.1  \\
Bidirectional iRNN (500 neurons, 5 layers)  &  28.9 / 29.7   \\\hline 
\end{tabular}
\caption{Frame error rates of recurrent methods on the TIMIT phone
  recognition task.}
\label{tab:speech}
\end{table}

In this task, instead of the identity initialization for the IRNNs
matrices we used $0.01 I$ so we refer to them as iRNNs.  Initalizing
with the full identity led to slow convergence, worse results and
sometimes led to the model diverging during training. We hypothesize
that this was because in the speech task similar inputs are provided
to the neural net in neighboring frames. The normal IRNN keeps
integrating this past input, instead of paying attention mainly to the
current input because it has a difficult time forgetting the past.  So
for the speech task, we are not only showing that iRNNs work much
better than RNNs composed of tanh units, but we are also showing that
initialization with the full identity is suboptimal when long range
effects are not needed. Mulitplying the identity with a small scalar
seems to be a good remedy in such cases.

In general in the speech recognition task, the iRNN easily outperforms
the RNN that uses tanh units and is comparable to LSTM although we
don't rule out the possibility that with very careful tuning of
hyperparameters, the relative performance of LSTMs or the iRNNs might
change.  A five layer Bidirectional LSTM outperforms all the other
models on this task, followed closely by a five layer Bidirectional
iRNN.


\subsection{Acknowledgements}
We thank Jeff Dean, Matthieu Devin, Rajat Monga, David Sussillo, Ilya
Sutskever and Oriol Vinyals for their help with the project.

\bibliography{translate}
\bibliographystyle{plain}
\end{document}